\documentclass{article} % For LaTeX2e
\usepackage{iclr2017_conference,times}
\usepackage{hyperref}
\usepackage{url}
\usepackage[pdftex]{graphicx}
\usepackage[utf8]{inputenc} % allow utf-8 input
\usepackage[T1]{fontenc}    % use 8-bit T1 fonts
\usepackage{booktabs}       % professional-quality tables
\usepackage{amsfonts}       % blackboard math symbols
\usepackage{nicefrac}       % compact symbols for 1/2, etc.
\usepackage{microtype}      % microtypography
\usepackage{graphicx}
\usepackage{pifont}

\title{Neural Photo Editing with Introspective Adversarial Networks}

\author{Andrew Brock, Theodore Lim, \& J.M. Ritchie\\
School of Engineering and Physical Sciences\\
Heriot-Watt University\\
Edinburgh, UK \\
\texttt{\{ajb5, t.lim, j.m.ritchie\}@hw.ac.uk} \\
\And
Nick Weston \\
Renishaw plc \\
Research Ave, North \\
Edinburgh, UK \\
\texttt{Nick.Weston@renishaw.com} \\
}

\newcommand{\subf}[2]{%
  {\small\begin{tabular}[b]{@{}c@{}}
  #1\\#2
  \end{tabular}}%
}

\iclrfinalcopy % Uncomment for camera-ready version
\definecolor{mygreen}{rgb}{0.032, 0.6392, 0.2039}

\newcommand{\cmark}{\textcolor{mygreen}{\ding{51}}}%
\newcommand{\xmark}{\textcolor{red}{\ding{55}}}%

\begin{document}

\maketitle

\begin{abstract}

The increasingly photorealistic sample quality of generative image models suggests their feasibility in applications beyond image generation. We present the Neural Photo Editor, an interface that leverages the power of generative neural networks to make large, semantically coherent changes to existing images. To tackle the challenge of achieving accurate reconstructions without loss of feature quality, we introduce the Introspective Adversarial Network,  
a novel hybridization of the VAE and GAN. Our model efficiently captures long-range dependencies through use of a computational block based on weight-shared dilated convolutions, and improves generalization performance with Orthogonal Regularization, a novel weight regularization method. We validate our contributions on CelebA, SVHN, and CIFAR-100, and produce samples and reconstructions with high visual fidelity.
\end{abstract}

\section{Introduction}
Editing photos typically involves some form of manipulating individual pixels, and achieving desirable results often requires significant user expertise. Given a sufficiently powerful image model, however, a user could quickly make large, photorealistic changes with ease by instead interacting with the model's controls. Two recent advances, the Variational Autoencoder (VAE)\citep{VAE} and Generative Adversarial Network (GAN)\citep{goodfellow2014generative}, have shown great promise for use in modeling the complex, high-dimensional distributions of natural images, but significant challenges remain before these models can be used as general-purpose image editors. 

VAEs are probabilistic graphical models that learn to maximize a variational lower bound on the likelihood of the data by projecting into a learned latent space, then reconstructing samples from that space. GANs learn a generative model by training one network, the "discriminator," to distinguish between real and generated data, while simultaneously training a second network, the "generator," to transform a noise vector into samples which the discriminator cannot distinguish from real data. Both approaches can be used to generate and interpolate between images by operating in a low-dimensional learned latent space, but each comes with its own set of benefits and drawbacks.

VAEs have stable training dynamics, but tend to produce images that discard high-frequency details when trained using maximum likelihood. Using the intermediate activations of a pre-trained discriminative neural network as features for comparing reconstructions to originals \citep{lamb2016discriminative} mollifies this effect, but requires labels in order to train the discriminative network in a supervised fashion.

By contrast, GANs have unstable and often oscillatory training dynamics, but produce images with sharp, photorealistic features. Basic GANs lack an inference mechanism, though techniques to train an inference network \citep{ALI} \citep{donahue2016adversarial} have recently been developed, as well as a hybridization that uses the VAE's inference network \citep{larsen2015autoencoding}.

Two key issues arise when attempting to use a latent-variable generative model to manipulate natural images. First, producing acceptable edits requires that the model be able to achieve close-to-exact reconstructions by inferring latents, or else the model's output will not match the original image. This simultaneously necessitates an inference mechanism (or inference-by-optimization) and careful design of the model architecture, as there is a tradeoff between reconstruction accuracy and learned feature quality that varies with the size of the information bottleneck.

Second, achieving a specific desired edit requires that the user be able to manipulate the model's latent variables in an interpretable way. Typically, this would require that the model's latent space be augmented during training and testing with a set of labeled attributes, such that interpolating along a latent such as "not smiling/smiling" produces a specific change. In the fully unsupervised setting, however, such semantically meaningful output features are generally controlled by an entangled set of latents which cannot be directly manipulated.

In this paper, we present the Neural Photo Editor, an interface that handles both of these issues, enabling a user to make large, coherent changes to the output of unsupervised generative models by indirectly manipulating the latent vector with a "contextual paintbrush." By applying a simple interpolating mask, we enable this same exploration for existing photos despite reconstruction errors.

Complementary to the Neural Photo Editor, we develop techniques to improve on common design tradeoffs in generative models. Our model, the Introspective Adversarial Network (IAN), is a hybridization of the VAE and GAN that leverages the power of the adversarial objective while maintaining the VAE's efficient inference mechanism, improving upon previous VAE/GAN hybrids both in parametric efficiency and output quality. We employ a novel convolutional block based on dilated convolutions \citep{dilated} to efficiently increase the network's receptive field, and Orthogonal Regularization, a novel weight regularizer.

We demonstrate the qualitative sampling, reconstructing, and interpolating ability of the IAN on CelebA \citep{liu2015deep}, SVHN \citep{netzer2011reading}, CIFAR-10 \citep{krizhevsky2009learning}, and Imagenet \citep{russakovsky2015imagenet}, and quantitatively demonstrate its inference capabilities with competitive performance on the semi-supervised SVHN classification task. Further quantitative experiments on CIFAR-100 \citep{krizhevsky2009learning} verify the generality of our dilated convolution blocks and Orthogonal Regularization.

\begin{figure}[tbp]
\begin{center}
{\includegraphics[scale=.4]{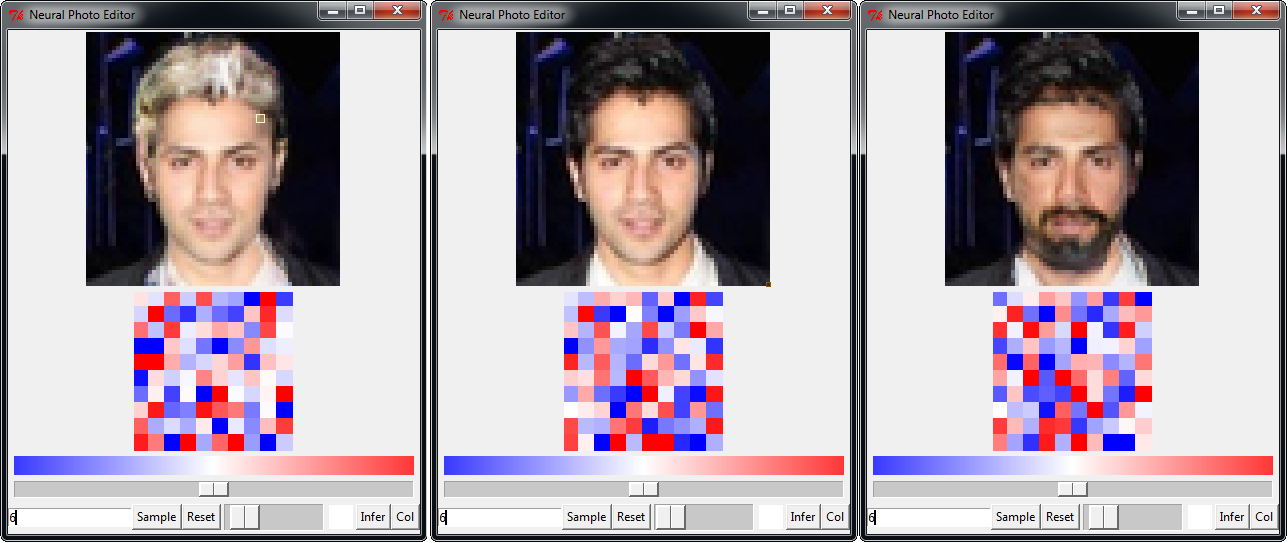}}
\end{center}
\caption{The Neural Photo Editor. The original image is center. The red and blue tiles are visualizations of the latent space, and can be directly manipulated as well.}
  \label{NPEVIS0}
\end{figure}

\section{Neural Photo Editing}
\label{NPE}

\begin{figure}[tbp]
\begin{center}
{\includegraphics[scale=.45]{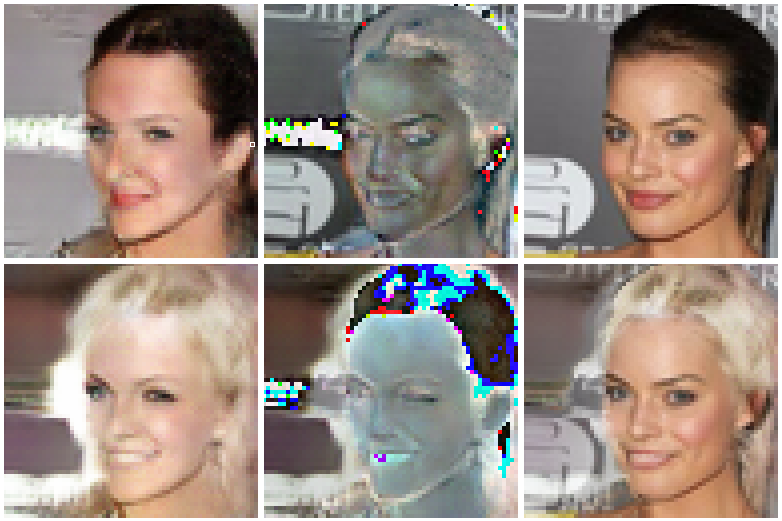}}
\end{center}
\caption{Visualizing the interpolation mask. Top, left to right: Reconstruction, reconstruction error, original image. Bottom: Modified reconstruction, $\Delta$, output.}
  \label{NPEVIS1}
\end{figure}

We present an interface, shown in Figure~\ref{NPEVIS0}, that turns a coarse user input into a refined, photorealistic image edit by indirectly manipulating the latent space with a "contextual paintbrush." The key idea is simple: a user selects a paint brush size and color (as with a typical image editor) and paints on the output image. Instead of changing individual pixels,  the interface backpropagates the difference between the local image patch and the requested color, and takes a gradient descent step in the latent space to minimize that difference. This step results in globally coherent changes that are semantically meaningful in the context of the requested color change. Given an output image $\hat{X}$ and a user requested color $X_{user}$, the change in latent values is $ -\frac{d||X_{user} - \hat{X}||_2}{dZ}$, evaluated at the current paintbrush location each time a user requests an edit.

For example, if a user has an image of a person with light skin, dark hair, and a widow's peak, by painting a dark color on the forehead, the system will automatically add hair in the requested area. Similarly, if a user has a photo of a person with a closed-mouth smile, the user can produce a toothy grin by painting bright white over the target's mouth.

This technique enables exploration of samples generated by the network, but fails when applied directly to existing photos, as it relies on the manipulated image being completely controlled by the latent variables, and reconstructions are usually imperfect. We circumvent this issue by introducing a simple masking technique that transfers edits from a reconstruction back to the original image. 

We take the output image to be a sum of the reconstruction, and a masked combination of the requested pixel-wise changes and the reconstruction error: 

\begin{equation}
Y = \hat{X} +M\Delta+(1-M)(X-\hat{X})
\end{equation}

Where $X$ is the original image, $\hat{X}$ is the model's reconstruction of $X$, and $\Delta$ is the difference between the modified reconstruction and $\hat{X}$. The mask $M$ is the channel-wise mean of the absolute value of $\Delta$, smoothed with a Gaussian filter $g$ and truncated pointwise to be between 0 and 1:
\begin{equation} M = min(g(\bar{|\Delta|}),1)
\end{equation}

The mask is designed to allow changes to the reconstruction to show through based on their magnitude. This relaxes the accuracy constraints by requiring that the reconstruction be feature-aligned rather than pixel-perfect, as only modifications to the reconstruction are applied to the original image. As long as the reconstruction is close enough and interpolations are smooth and plausible, the system will successfully transfer edits.

A visualization of the masking technique is shown in Figure~\ref{NPEVIS1}. This method adds minimal computational cost to the underlying latent space exploration and produces convincing changes of features including hair color and style, skin tone, and facial expression. A video of the interface in action is available online.\footnote{https://www.youtube.com/watch?v=FDELBFSeqQs}

\section{Introspective Adversarial Networks}
\label{IAN}

\begin{figure}[tbp]
\begin{center}
{\includegraphics[scale=.45]{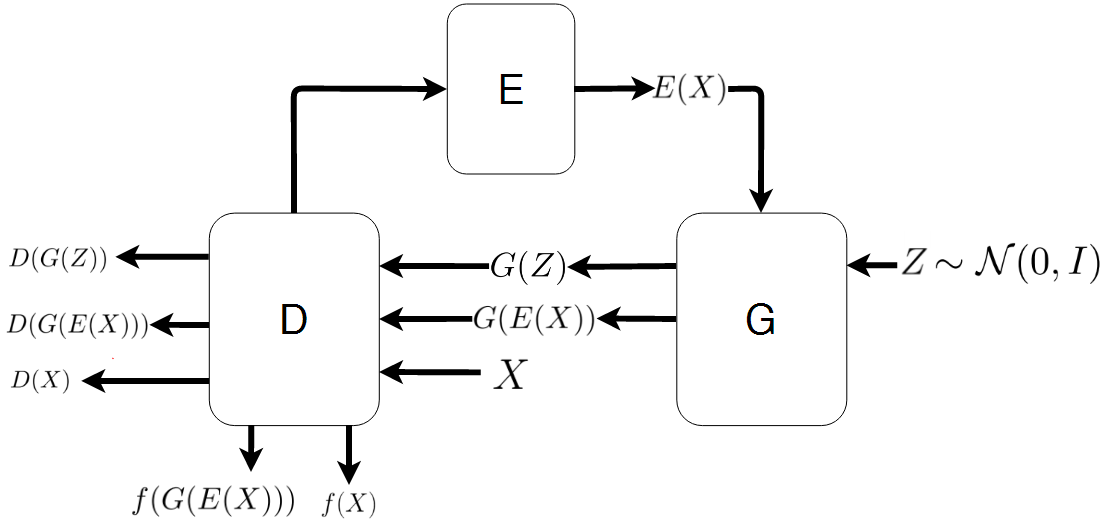}}
\end{center}
\caption{The Introspective Adversarial Network (IAN).}
  \label{IANDIAG}
\end{figure}

Complementary to the Neural Photo Editor, we introduce the Introspective Adversarial Network (IAN), a novel hybridization of the VAE and GAN motivated by the need for an image model with photorealistic outputs that  achieves high-quality reconstructions without loss of representational power. There is typically a design tradeoff between these two goals related to the size of the latent space: a higher-dimensional latent space (i.e. a wider representational bottleneck) tends to learn less descriptive features, but produces higher quality reconstructions.

We thus seek techniques to improve the capacity of the latent space without increasing its dimensionality. Similar to VAE/GAN \citep{larsen2015autoencoding}, we use the decoder network of the autoencoder as the generator network of the GAN, but instead of training a separate discriminator network, we combine the encoder and discriminator into a single network. Central to the IAN is the idea that features learned by a discriminatively trained network tend to be more expressive those learned by an encoder network trained via maximum likelihood (i.e. more useful on semi-supervised tasks), and thus better suited for inference. As the Neural Photo Editor relies on high-quality reconstructions, the inference capacity of the underlying model is critical. Accordingly, we use the discriminator of the GAN, $D$, as a feature extractor for an inference subnetwork, $E$, which is implemented as a fully-connected layer on top of the final convolutional layer of the discriminator. We infer latent values $Z \sim E(X) = q(Z|X)$ for reconstruction and sample random values $Z \sim p(Z)$ from a standard normal for random image generation using the generator network, $G$.

Similar to VAE/GAN and DeePSiM \citep{Brox}, we use three distinct loss functions:
\begin{itemize}

\item $ \mathcal{L}_{img}$, the $ \mathcal{L}_{1}$ pixel-wise reconstruction loss, which we prefer to the $ \mathcal{L}_{2}$ reconstruction loss for its higher average gradient.

\item $ \mathcal{L}_{feature}$, the feature-wise reconstruction loss, evaluated as the $ \mathcal{L}_{2}$ difference between the original and reconstruction in the space of the hidden layers of the discriminator.

\item $ \mathcal{L}_{adv}$, the ternary adversarial loss, a modification of the adversarial loss that forces the discriminator to label a sample as real, generated, or reconstructed (as opposed to a binary real vs. generated label).

\end{itemize}

Including the VAE's KL divergence between the inferred latents $E(X)$ and the prior $p(Z)$, the loss function for the generator and encoder network is thus:

\begin{equation}
\mathcal{L}_{E,G} = \lambda_{adv}\mathcal{L}_{Gadv} + \lambda_{img}\mathcal{L}_{img} + \lambda_{feature}\mathcal{L}_{feature} + D_{KL}(E(X)||p(Z))
\end{equation}

Where the $\lambda$ terms weight the relative importance of each loss. We set $\lambda_{img}$ to 3 and leave the other terms at 1. The discriminator is updated solely using the ternary adversarial loss. During each training step, the generator produces reconstructions $G(E(X))$ (using the standard VAE reparameterization trick) from data $X$ and random samples $G(Z)$, while the discriminator observes $X$ as well as the reconstructions and random samples, and both networks are simultaneously updated.

\subsection{Feature-wise Loss}
We compare reconstructions using the intermediate activations, $f(G(E(X)))$, of all convolutional layers of the discriminator, mirroring the perceptual losses of  Discriminative Regularization \citep{lamb2016discriminative}, VAE/GAN \citep{larsen2015autoencoding}, and DeepSiM \citep{Brox}. We note that Feature Matching \citep{ImprovedGAN} is designed to operate in a similar fashion, but without the guidance of an inference mechanism to match latent values $Z$ to particular values of $f(G(Z))$. We find that using this loss to complement the pixel-wise difference results in sharper reconstructions that better preserve higher frequency features and edges.

\subsection{Ternary Adversarial Loss}
The standard GAN discriminator network is trained using an implicit label source (real vs fake); noting the success of augmenting the discriminator's objective with supervised labels \citep{ACGAN}, we seek additional sources of implicit labels, in the hopes of achieving similar improvements. The ternary loss provides an additional source of supervision to the discriminator by asking it to determine if a sample is real, generated, or a reconstruction, while the generator's goal is still to have the discriminator assign a high "real" probability to both samples and reconstructions. We thus modify the discriminator to have three output units with a softmax nonlinearity, and train it to minimize the categorical cross-entropy:

\begin{equation}\label{eq:4} \mathcal{L}_{Dadv} = -log(D_{real}(X)) - log(D_{generated}(G(Z))) - log(D_{reconstructed}(G(E(X)))) 
 \end{equation}

Where each $D$ term in Equation ~\ref{eq:4} indicates the discriminator output unit assigned to each label class. 
The generator is trained to produce outputs that maximize the probability of the label "real" being assigned by the discriminator by minimizing $\mathcal{L}_{Gadv}$:

\begin{equation} \mathcal{L}_{Gadv} = -log(D_{real}(G(Z))) - log(D_{real}(G(E(X))) 
\end{equation}

We posit that this loss helps maintain the balance of power early in training by preventing the discriminator from learning a small subset of features (e.g. artifacts in the generator's output) that distinguish real and generated samples, reducing the range of useful features the generator can learn from the discriminator. We also find that this loss leads to higher sample quality, perhaps because the additional source of supervision leads to the discriminator ultimately learning a richer feature space.

\subsection{Architecture}

Our model has the same basic structure as DCGAN \citep{radford2015unsupervised}, augmented with  Multiscale Dilated Convolution (MDC) blocks in the generator, and Minibatch Discrimination \citep{ImprovedGAN} in the discriminator. As in \citep{radford2015unsupervised}, we use Batch Normalization \citep{Bnorm} and Adam \citep{Adam} in both networks. All of our code is publicly available.\footnote{https://github.com/ajbrock/Neural-Photo-Editor}

\subsection{Multiscale Dilated Convolution Blocks}

\begin{figure}[tbp]
  \centering
  \begin{tabular}{cc}
  \subf{\includegraphics[scale=.45]{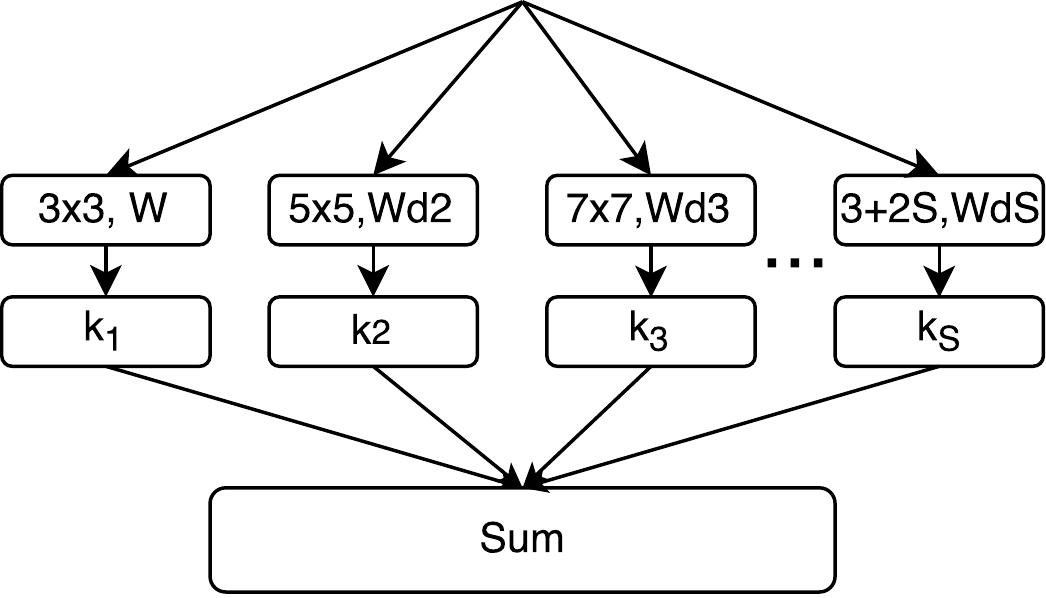}}{(a)}

&
 \subf{\includegraphics[scale=0.45]{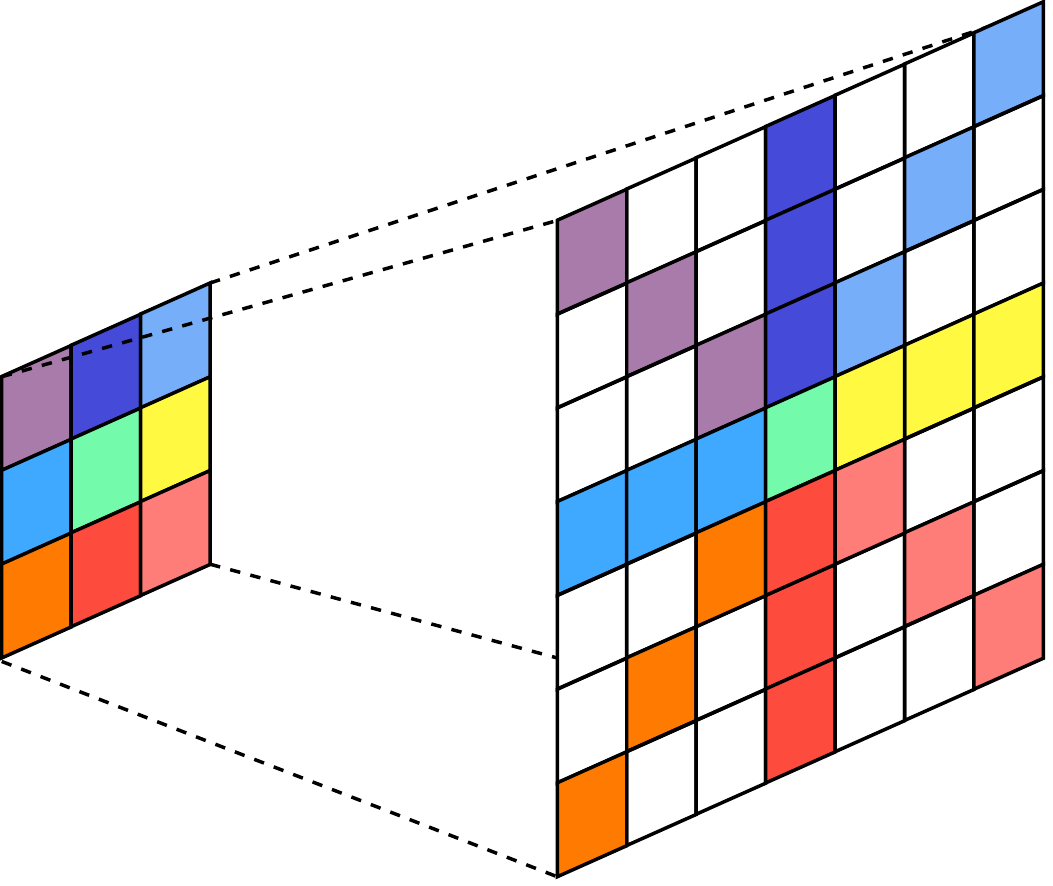}}{(b)}
  
 \end{tabular}
 \caption{(a) Multiscale Dilated Convolution Block. (b) Visualizing a 3d3 MDC filter composition.}
 \label{MDCBlock} 
\end{figure}

We propose a novel Inception-style \citep{Inception} convolutional block motivated by the ideas that image features naturally occur at multiple scales, that a network’s expressivity is proportional to the range of functions it can represent divided by its total number of parameters, and by the desire to efficiently expand a network's receptive field. The Multiscale Dilated Convolution (MDC) block applies a single FxF filter at multiple dilation factors, then performs a weighted elementwise sum of each dilated filter’s output, allowing the network to simultaneously learn a set of features and the relevant scales at which those features occur with a minimal increase in parameters. This also rapidly expands the network's receptive field without requiring an increase in depth or the number of parameters. Dilated convolutions have previously been successfully applied in semantic segmentation \citep{dilated}, and a similar scheme, minus the parameter sharing, is proposed in \citep{AtrousCRF}.

As shown in Figure~\ref{MDCBlock}(a), each block is parameterized by a bank of N FxF filters $W$, applied with S factors of dilation, and a set of N*S scalars $k$, which relatively weight the output of each filter at each scale. This is naturally and efficiently implemented by reparameterizing a sparsely populated F+(S-1)*(F-1) filterbank  as displayed in Figure~\ref{MDCBlock}(b). We propose two variants: Standard MDC, where the filter weights are tied to a base $W$, and Full-Rank MDC, where filters are given the sparse layout of Figure~\ref{MDCBlock}(b) but the weights are not tied. Selecting Standard versus Full-Rank MDC blocks allows for a design tradeoff between parametric efficiency and model flexibility. In our architecture, we replace the hidden layers of the generator with Standard MDC blocks, using F=5 and D=2; we specify MDC blocks by their base filter size and their maximum dilation factor (e.g. 5d2).

\subsection{Orthogonal Regularization}

Orthogonality is a desirable quality in ConvNet filters, partially because multiplication by an orthogonal matrix leaves the norm of the original matrix unchanged. This property is valuable in deep or recurrent networks, where repeated matrix multiplication can result in signals vanishing or exploding. We note the success of initializing weights with orthogonal matrices \citep{Orthog}, and posit that maintaining orthogonality throughout training is also desirable. To this end, we propose a simple weight regularization technique, Orthogonal Regularization, that encourages weights to be orthogonal by pushing them towards the nearest orthogonal manifold. We augment our objective with the cost: 

\begin{equation}\mathcal{L}_{ortho} = \Sigma(| WW^{T} - I|)
\end{equation}

Where $\Sigma$ indicates a sum across all filter banks, $W$ is a filter bank, and $I$ is the identity matrix. 

\section{Related Work}
\label{related_work}

Our architecture builds directly off of previous VAE/GAN hybrids \citep{larsen2015autoencoding} \citep{Brox}, with the key difference being our combination of the discriminator and the encoder to improve computational and parametric efficiency (by reusing discriminator features) as well as reconstruction accuracy (as demonstrated in our CelebA ablation studies). The methods of ALI \citep{ALI} and BiGAN \citep{donahue2016adversarial} provide an orthogonal approach to GAN inference, in which an inference network is trained by an adversarial (as opposed to a variational) process.

The method of iGAN \citep{GVM} bears the most relation to our interface. The iGAN interface allows a user to impose shape or color constraints on an image of an object through use of a brush tool, then optimizes to solve for the output of a DCGAN \citep{radford2015unsupervised} which best satisfies those constraints. Photorealistic edits are transferred to existing images via motion and color flow estimation.

Both iGAN and the Neural Photo Editor turn coarse user input into refined outputs through use of a generative model, but the methods differ in several key ways. First, we focus on editing portraits, rather than objects such as shoes or handbags, and are thus more concerned with modifying \textit{features}, as opposed to overall color or shape, for which our method is less well-suited. Our edit transfer technique follows this difference as well: we directly transfer the local image changes produced by the model back onto the original image, rather than estimating and mimicking motion and color flow.

Second, our interface applies user edits one step at a time, rather than iteratively optimizing the output. This highlights the difference in design approaches: iGAN seeks to produce outputs that best match a given set of user constraints, while we seek to allow a user to guide the latent space traversal.

Finally, we explicitly tailor our model design to the task at hand and jointly train an inference network which we use at test time to produce reconstructions in a single shot. In contrast, iGAN trains an inference network to minimize the $\mathcal{L}_2$ loss after training the generator network, and use the inference network to get an initial estimate of the inferred latents, which are then iteratively optimized.

Another related interface \citep{neuraldoodle} refines simple user input into complex textures through use of artistic style transfer \citep{StyleTransfer}.
Other related work \citep{TomWhite} also circumvents the need for labeled attributes by constructing latent vectors by analogy and bias-correcting them.

\section{Experiments}
\begin{figure}[tbp]
\begin{center}
{\includegraphics[scale=.375]{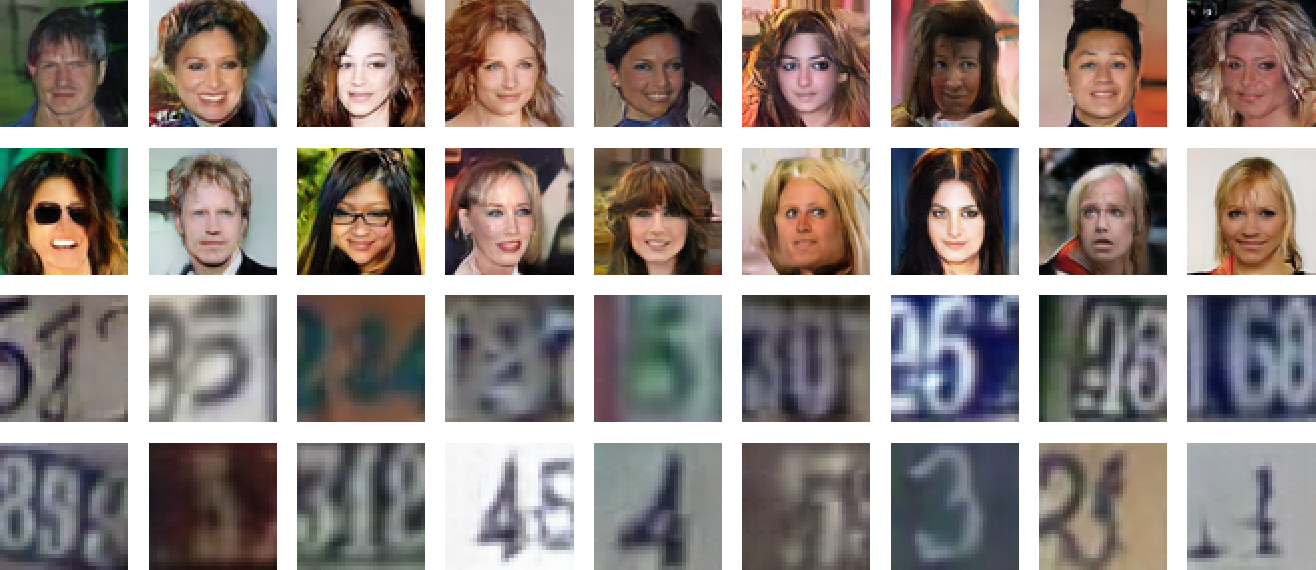}}
\end{center}
\caption{CelebA  and SVHN samples.}
  \label{samples}
\end{figure}
\begin{figure}[tbp]
\begin{center}
{\includegraphics[scale=.375]{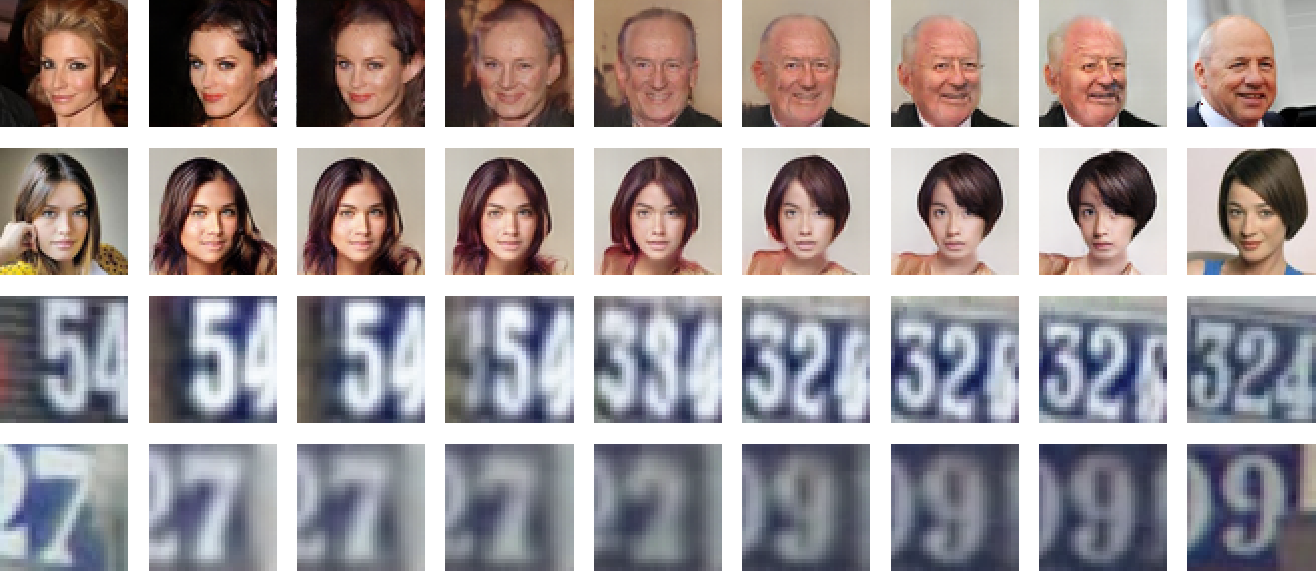}}
\end{center}
\caption{CelebA  and SVHN Reconstructions and Interpolations. The outermost images are originals, the adjacent images are reconstructions.}
  \label{recon}
\end{figure}

We qualitatively evaluate the IAN on 64x64 CelebA \citep{liu2015deep},  32x32 SVHN \citep{netzer2011reading}, 32x32 CIFAR-10 \citep{krizhevsky2009learning}, and 64x64 Imagenet \citep{russakovsky2015imagenet}.  Our models are implemented in Theano \citep{Theano} with Lasagne \citep{Lasagne}. Samples from the IAN, randomly selected and shown in Figure~\ref{samples}, display the visual fidelity typical of adversarially trained networks.  The IAN demonstrates high quality reconstructions on previously unseen data, shown in Figure~\ref{recon}, and smooth, plausible interpolations, even between drastically different samples. CIFAR and Imagenet samples, along with additional comparisons to samples from other models, are available in the appendix.

\subsection{Discriminative Experiments}

We quantitatively demonstrate the effectiveness of our MDC blocks and Orthogonal Regularization on the CIFAR-100 \citep{krizhevsky2009learning} benchmark. Using standard data augmentation, we train a set of 40-layer, k=12 DenseNets \citep{DenseNets} for 50 epochs, annealing the learning rate at 25 and 37 epochs. We add varying amounts of Orthogonal Regularization and modify the standard DenseNet architecture by replacing every 3x3 filterbank with 3d3 MDC blocks, and report the test error after training in Table~\ref{cifartable}. In addition, we compare to performance using full 7x7 filters.

%In addition, we run a 40-Layer, k=12 DenseNet using "full" MDC blocks (where each filter has the same layout of a 3d2 MDC block, but where all parameters are free rather than tied to a base 3x3 filter) for 300 epochs.

There is a noticeable increase in performance with the progressive addition of our modifications, despite a negligible increase in the number of parameters.
Adding Orthogonal Regularization improves the network's generalization ability; we suspect this is because it encourages the filter weights to remain close to a desirable, non-zero manifold, increasing the likelihood that all of the available model capacity is used by preventing the magnitude of the weights from overly diminishing.   Replacing 3x3 filters with MDC blocks yields additional performance gains; we suspect this is due to an increase in the expressive power and receptive field of the network, allowing it to learn longer-range dependencies with ease. We also note that substituting Full-Rank MDC blocks into a 40-Layer DenseNet improves performance by a relative 5\%, with the only increased computational cost coming from using the larger filters.

For use in evaluating the IAN, we additionally train 40-layer, k=12 DenseNets on the CelebA attribute classification task with varying amounts of Orthogonal Regularization. A plot of the train and validation error during training is available in Figure 7. The addition of of Orthogonal Regularization improves the validation error from 6.55\% to 4.22\%, further demonstrating its utility.

\subsection{Evaluating Modifications}
For use in editing photos, a model must produce reconstructions which are photorealistic and feature-aligned, and have smooth, plausible interpolations between outputs. We perform an ablation study to investigate the effects  of our proposals, and employ several metrics to evaluate model quality given these goals. In this study, we progressively add modifications to a VAE/GAN \citep{larsen2015autoencoding} baseline, and train each network for 50 epochs.

\begin{table}[tbp]
\begin{center}\begin{tabular}{c|c|c|c|c} 
\hline
Model & \# Params &  MDC & Ortho. Reg. & Error (\%) \\
\hline Baseline DenseNet (D=40,K=12) & 1.0M & \xmark & \xmark & 26.71 \\
\hline DenseNet with Ortho. Reg. & 1.0M & \xmark & 1e-3 & 26.51 \\
\hline DenseNet with Ortho. Reg & 1.0M & \xmark & 1e-1 & 26.46 \\
\hline DenseNet with 7x7 Filters & 5.0M &\xmark & \xmark & 26.39\\
\hline DenseNet with 3d3 MDC & 1.0M &\cmark & \xmark & 26.02\\
\hline DenseNet with Ortho. Reg \& MDC & 1.0M &\cmark & 1e-3 & 25.72 \\
\hline DenseNet with Ortho. Reg \& MDC & 1.0M & \cmark & 1e-1 & 25.39 \\ 
\hline DenseNet \citep{DenseNets}, 300 epochs & 1.0M & \xmark & \xmark &24.42 \\
\hline {\bf DenseNet with Full MDC, 300 epochs} & 2.8M & \bf{full} & \bf{\xmark} & {\bf 23.30} \\
\hline

\end{tabular}
\end{center}
\caption{\label{cifartable}Error rates on CIFAR-100+ after 50 epochs.}
\end{table}

For reconstruction accuracy, pixel-wise distance does not tend to correlate well with perceptual similarity. In addition to pixel-wise $\mathcal{L}_2$ distance, we therefore compare model reconstruction accuracy in terms of:
\begin{itemize}
\item Feature-wise $\mathcal{L}_2$ distance in the final layer of a 40-Layer k=12 DenseNet trained for the CelebA attribute classification task.

\item Trait reconstruction error. We run our classification DenseNet to predict a  binary attribute vector $y(X)$ given an image X, and $y(G(E(X)))$ given a model's reconstruction, then measure the percent error.

\item Fiducial keypoint Error, measured as the mean $\mathcal{L}_2$ distance between the facial landmarks predicted by the system of \citep{umdfaces-paper}.

\end{itemize}

Gauging the visual quality of the model's outputs is notoriously difficult, but the Inception score recently proposed by \citep{ImprovedGAN} has been found to correlate positively with human-evaluated sample quality. Using our CelebA attribute classification network in place of the Inception \citep{Inception} model, we compare  the Inception score of each model  evaluated on 50,000 random samples. We posit that this metric is also indicative of interpolation quality, as a high visual quality score on a large sample population suggests that the model's output quality remains high regardless of the state of the latent space. 

Results of this ablation study are presented in Table~\ref{celebtable}; samples and reconstructions from each configuration are available in the appendix, along with comparisons between a fully-trained IAN and related models. As with our discriminative experiments, we find that the progressive addition of modifications results in consistent performance improvements across our reconstruction metrics and the Inception score.

We note that the single largest gains come from the inclusion of MDC blocks, suggesting that the network's receptive field is a critical aspect of network design for both generative and discriminative tasks, with an increased receptive field correlating positively with reconstruction accuracy and sample quality.

The improvements from Orthogonal Regularization suggest that encouraging weights to lie close to the orthogonal manifold is beneficial for improving the sample and reconstruction quality of generative neural networks by preventing learned weights from collapsing to an undesirable manifold; this is consistent with our experience iterating through network designs, where we have found mode collapse to occur less frequently while using Orthogonal Regularization.

Finally, the increase in sample quality and reconstruction accuracy through use of the ternary adversarial loss suggests that including the "reconstructed" target in the discriminator's objective does lead to the discriminator learning a richer feature space. This comes along with our observations that training with the ternary loss, where we have observed that the generator and discriminator losses tend to be more balanced than when training with the standard binary loss.

%--our inception scores are summed across 40 logistic distributions rather than a single softmax and therefore operate on larger scales.
%--We further verify the discriminative quality of this metric by evaluating the inception score of a mode-collapsed DCGAN, which produces an inception score of 

\begin{table}[tbp]
\begin{center}\begin{tabular}{c|c|c|c|c|c|c|c} 
\hline
 MDC & Ortho. Reg. & Ternary & Pixel & Feature & Trait(\%) & Keypoint  & Inception \\
 \hline \multicolumn{3}{c|}{VAE/GAN Baseline} &0.295 & 4.86 &0.197 & 2.21 & $1389 (\pm 64)$\\
 
\hline   \xmark & \xmark &\xmark & 0.285& 4.76 & 0.189 & 2.11 & $1772 (\pm 37)$\\

\hline   \xmark & \cmark & \xmark & 0.258 & 4.67 & 0.182 & 1.79 & $2160 (\pm 70)$ \\

\hline   \cmark & \xmark & \xmark &  0.248 & 4.69 & 0.172 & 1.54  & $2365 (\pm 97)$\\

\hline    \cmark & \cmark & \xmark &0.230 & 4.39 & 0.165 & 1.47 & $3158 (\pm 98)$\\

\hline   \xmark & \xmark & \cmark &0.254 & 4.60 & 0.177 & 1.67 & $2648 (\pm 69)$\\

\hline    \xmark & \cmark & \cmark & 0.239 & 4.51 & 0.164 & 1.57  & $3161 (\pm 70)$\\

\hline   \cmark & \xmark & \cmark &  0.221 &  4.37 & 0.158 & 0.99 &$3300 (\pm 123)$\\
 
\hline  \cmark & \cmark & \cmark & 0.192 & 4.33 & 0.155 & 0.97 &$3627 (\pm 146)$\\

\hline

\end{tabular}
\end{center}

\caption{\label{celebtable} CelebA investigations.}
\end{table}

\subsection{Semi-Supervised learning with SVHN}

\begin{table}[tbp]
\begin{tabular}{cc}

\subf{
\begin{tabular}{ll} \toprule
Method                                         & Error rate                      \\ \midrule
VAE (M1 + M2) \citep{kingma2014semi}           & $36.02\%$
\\
SWWAE with dropout \citep{zhao2015stacked}     & $23.56\%$                       \\
DCGAN + L2-SVM \citep{radford2015unsupervised} & $22.18\% (\pm 1.13\%)$          \\
 SDGM \citep{maaloe2016auxiliary}         &  $16.61\% (\pm 0.24\%)$ \\
 
ALI (L2-SVM) \citep{ALI}                                & $19.14\% (\pm 0.50\%)$          \\ 
IAN (ours, L2-SVM)                                     & $18.50\% (\pm 0.38\%)$ \\

 \midrule
IAN (ours, Improved-GAN)                                     & $8.34\% (\pm 0.91\%)$ \\

 Improved-GAN \citep{ImprovedGAN}    &  $8.11\% (\pm 1.3\%)$\\

{\bf ALI (Improved-GAN)} & $\mathbf{7.3\%}$          \\ 
\hline

\end{tabular}}{Table 3}

\subf{\includegraphics[scale=0.35]{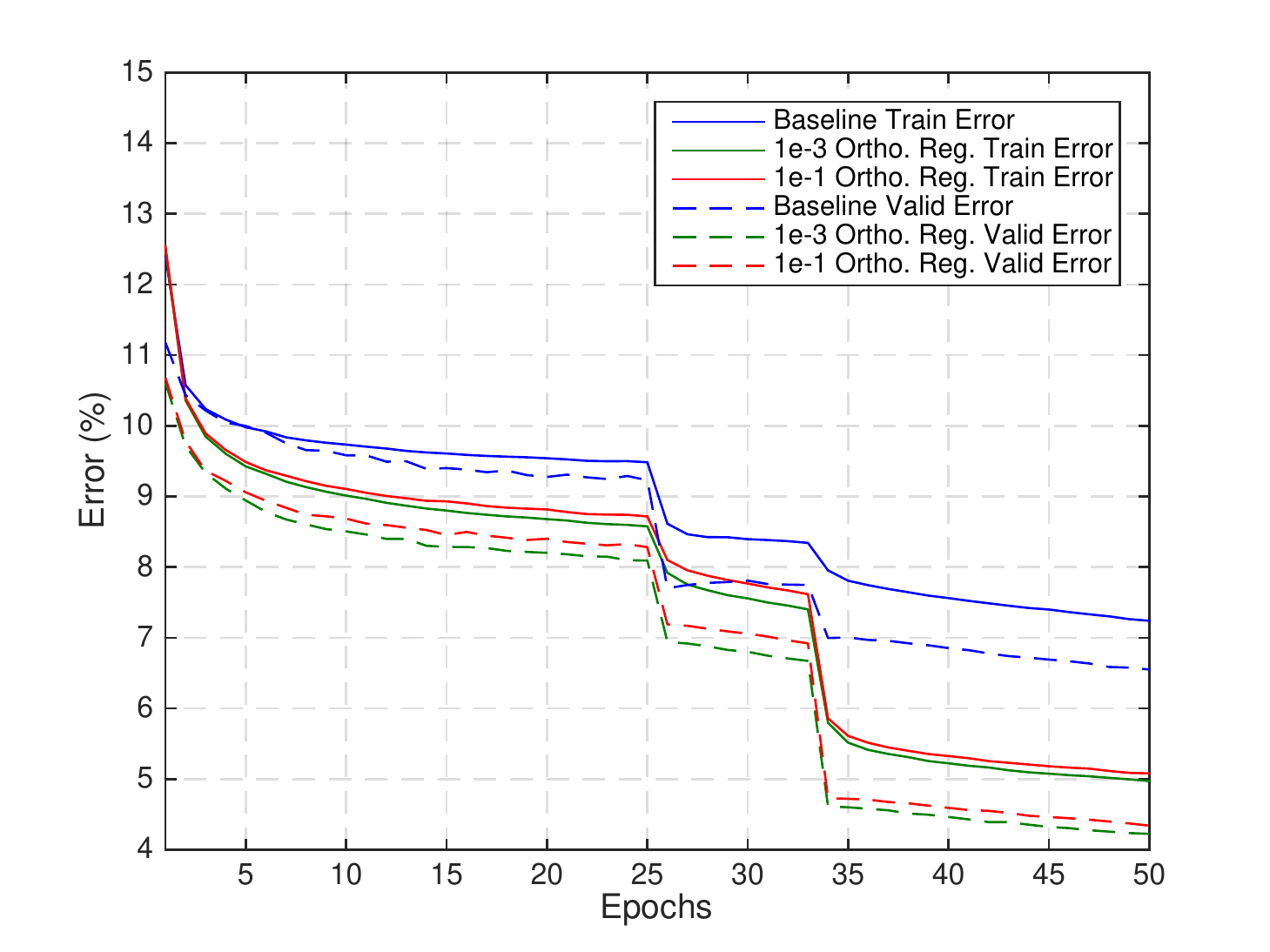}}{Figure 7}

\end{tabular}
\caption{\label{semitable}  Error rates on Semi-Supervised SVHN with 1000 training examples. Figure 7: Performance on CelebA Classification task with varying Orthogonal Regularization. }
\end{table}

We quantitatively evaluate the inference abilities of our architecture by applying it to the semi-supervised SVHN classification task using two different procedures. We first evaluate using the procedure of \citep{radford2015unsupervised} by training an L2-SVM on the output of the FC layer of the encoder subnetwork, and report average test error and standard deviation across 100 different SVMs, each trained on 1000 random examples from the training set.

Next, we use the procedure of \citep{ImprovedGAN}, where the discriminator outputs a distribution over the $K$ object categories and an additional "fake" category, for a total of $K$+1 outputs. The discriminator is trained to predict the category when given labeled data, to assign the "fake" label when provided data from the generator, and to assign $k \in \{1,...,K\}$ when provided unlabeled real data. We modify feature-matching based Improved-GAN to include the encoder subnetwork and reconstruction losses detailed in Section~\ref{IAN}, but do not include the ternary adversarial loss.

Our performance, as shown in Table~\ref{semitable}, is competitive with other networks evaluated in these fashions, achieving 18.5\% mean classification accuracy when using SVMs and 8.34\% accuracy when using the method of Improved-GAN. When using SVMs, our method tends to demonstrate improvement over previous methods, particularly over standard VAEs. We believe this is due to the encoder subnetwork being based on more descriptive features (i.e. those of the discriminator), and therefore better suited to discriminating between SVHN classes.

We find the lack of improvement when using the method of Improved-GAN unsurprising, as the IAN architecture does not change the goal of the discriminator; any changes in behavior are thus indirectly due to changes in the generator, whose loss is only slightly modified from feature-matching Improved-GAN.

\section{Conclusion}
We introduced the Neural Photo Editor, a novel interface for exploring the learned latent space of generative models and for making specific semantic changes to natural images. Our interface makes use of the Introspective Adversarial Network, a hybridization of the VAE and GAN that outputs high fidelity samples and reconstructions, and achieves competitive performance in a semi-supervised classification task. The IAN makes use of Multiscale Dilated Convolution Blocks and Orthogonal Regularization, two improvements designed to improve model expressivity and feature quality for convolutional networks.

\subsubsection*{Acknowledgments}

This research was made possible by grants and support from Renishaw plc and the Edinburgh Centre For Robotics. The work presented herein is also partially funded under the European H2020 Programme BEACONING project, Grant Agreement nr. 687676.

\bibliography{IAN}
\bibliographystyle{iclr2017_conference}

\newpage

\section*{Appendix: Additional Visual Comparisons}
\begin{figure}[htbp]
\begin{center}
{\includegraphics[scale=0.62]{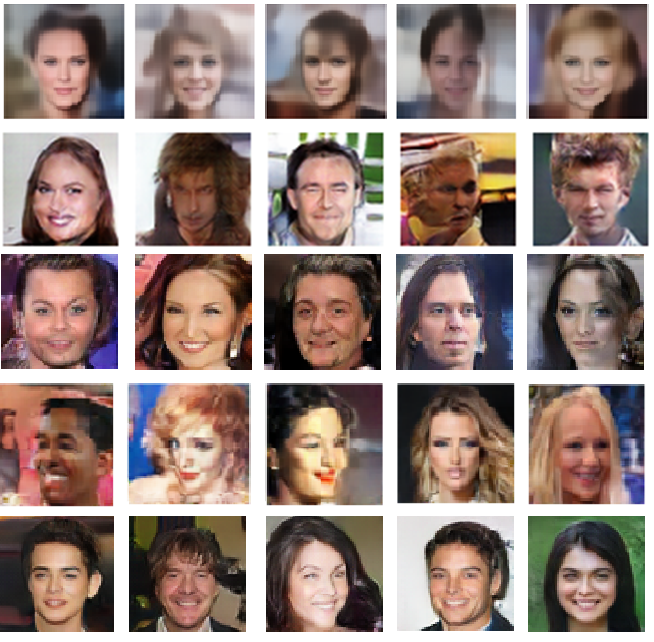}}
\end{center}
\caption{Comparing samples from different models. From top: VAE\citep{VAE}, DCGAN \citep{goodfellow2014generative}, VAE/GAN from \citep{larsen2015autoencoding}, ALI from\citep{ALI}, IAN (ours).}
\label{DCGAN_VAE_IAN}
\end{figure}

\begin{table}[tbp]
\begin{center}\begin{tabular}{c|c|c|ccccc} 
\hline
 MDC & Ortho. Reg. & Ternary & Recon1 & Recon2 & Sample1 & Sample 2 & Sample 3 \\
  \hline \multicolumn{3}{c|}{Original} &{\includegraphics[scale=0.375]{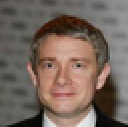}} &  {\includegraphics[scale=0.375]{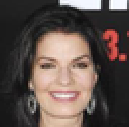}} &&& \\

 \multicolumn{3}{c|}{VAE/GAN Baseline} &{\includegraphics[scale=0.375]{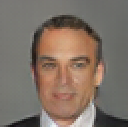}} &  {\includegraphics[scale=0.375]{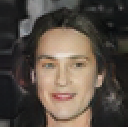}} &
  {\includegraphics[scale=0.375]{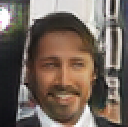}} &
  {\includegraphics[scale=0.375]{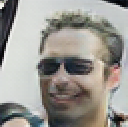}} &
  {\includegraphics[scale=0.375]{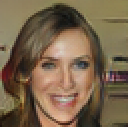}} \\

 \hline  \xmark & \xmark &\xmark 
&{\includegraphics[scale=0.375]{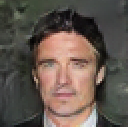}} &  {\includegraphics[scale=0.375]{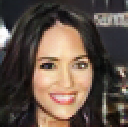}} &
  {\includegraphics[scale=0.375]{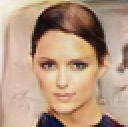}} &
  {\includegraphics[scale=0.375]{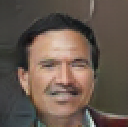}} &
  {\includegraphics[scale=0.375]{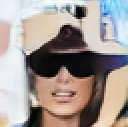}} \\

   \xmark & \cmark & \xmark 
&{\includegraphics[scale=0.375]{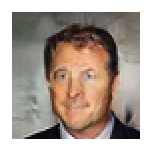}} &  {\includegraphics[scale=0.375]{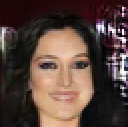}} &
  {\includegraphics[scale=0.375]{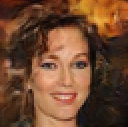}} &
  {\includegraphics[scale=0.375]{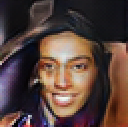}} &
  {\includegraphics[scale=0.375]{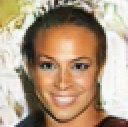}} \\

  \cmark & \xmark & \xmark
&{\includegraphics[scale=0.375]{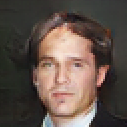}} &  {\includegraphics[scale=0.375]{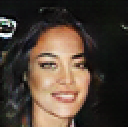}} &
  {\includegraphics[scale=0.375]{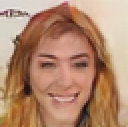}} &
  {\includegraphics[scale=0.375]{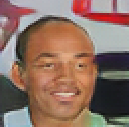}} &
  {\includegraphics[scale=0.375]{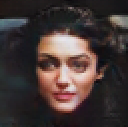}} \\

   \cmark & \cmark & \xmark 
&{\includegraphics[scale=0.375]{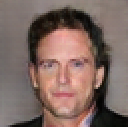}} &  {\includegraphics[scale=0.375]{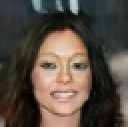}} &
  {\includegraphics[scale=0.375]{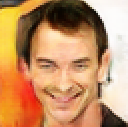}} &
  {\includegraphics[scale=0.375]{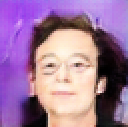}} &
  {\includegraphics[scale=0.375]{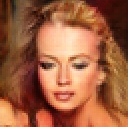}} \\

   \xmark & \xmark & \cmark 
&{\includegraphics[scale=0.375]{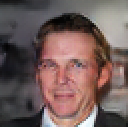}} &   		   {\includegraphics[scale=0.375]{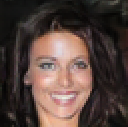}} &
  {\includegraphics[scale=0.375]{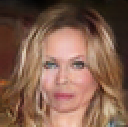}} &
  {\includegraphics[scale=0.375]{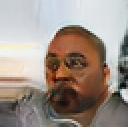}} &
  {\includegraphics[scale=0.375]{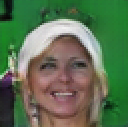}} \\

    \xmark & \cmark & \cmark 
&{\includegraphics[scale=0.375]{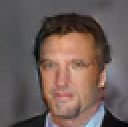}} &   		   {\includegraphics[scale=0.375]{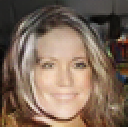}} &
  {\includegraphics[scale=0.375]{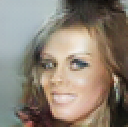}} &
  {\includegraphics[scale=0.375]{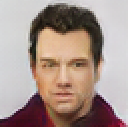}} &
  {\includegraphics[scale=0.375]{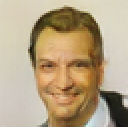}} \\

   \cmark & \xmark & \cmark 
&{\includegraphics[scale=0.375]{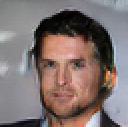}} &   		   {\includegraphics[scale=0.375]{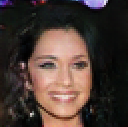}} &
  {\includegraphics[scale=0.375]{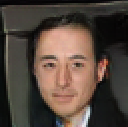}} &
  {\includegraphics[scale=0.375]{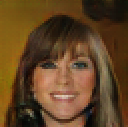}} &
  {\includegraphics[scale=0.375]{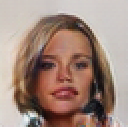}} \\
 
  \cmark & \cmark & \cmark 
&{\includegraphics[scale=0.375]{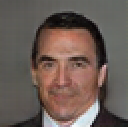}} &   		   {\includegraphics[scale=0.375]{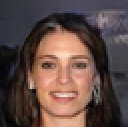}} &
  {\includegraphics[scale=0.375]{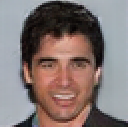}} &
  {\includegraphics[scale=0.375]{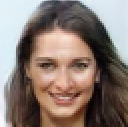}} &
  {\includegraphics[scale=0.375]{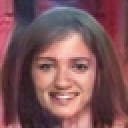}} \\

\hline

\end{tabular}
\end{center}

\caption{\label{celebablationpics} Reconstructions and samples from CelebA ablation Study.}
\end{table}

\begin{figure}[htbp]
\begin{center}
{\includegraphics[scale=0.85]{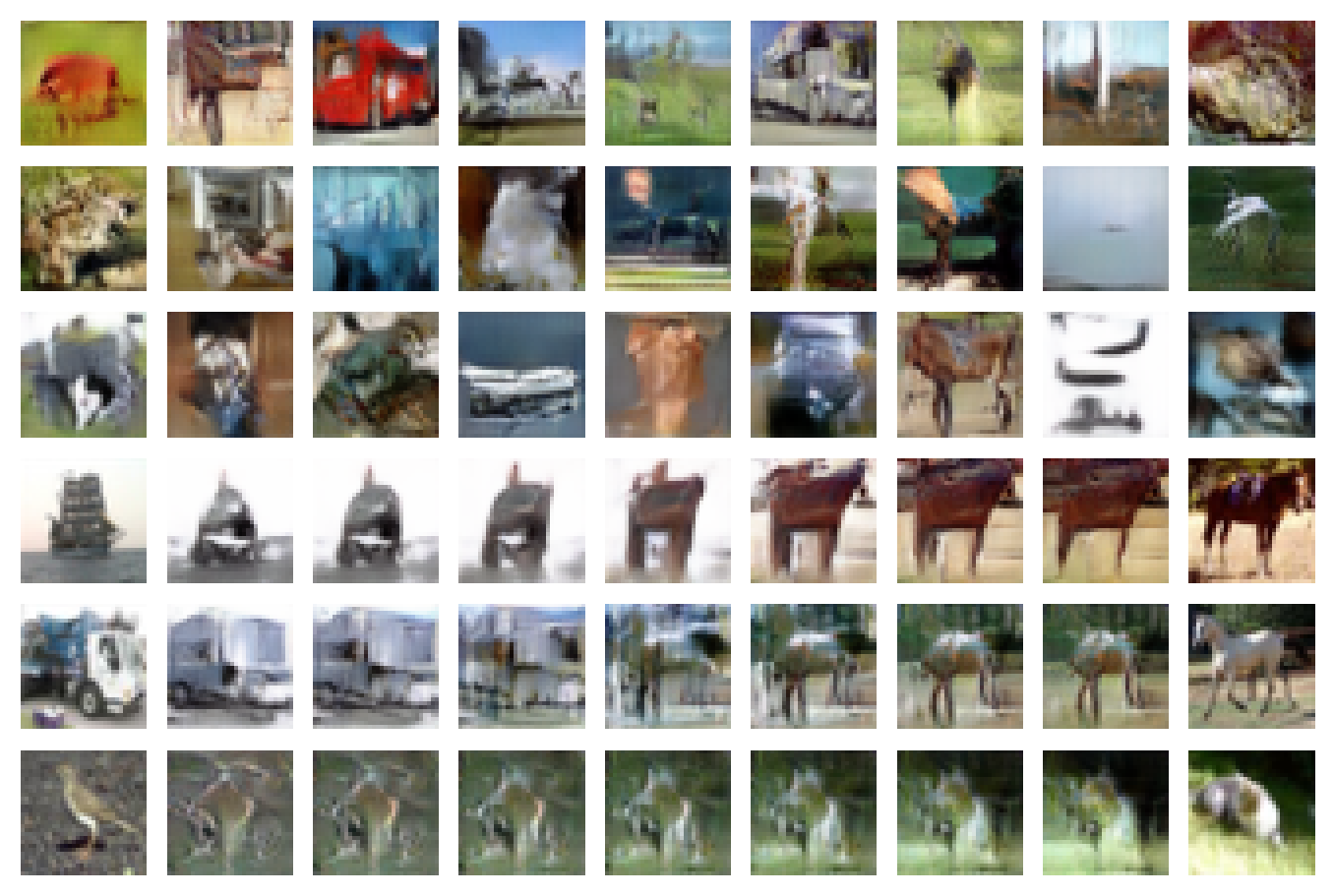}}
\end{center}
\caption{Samples, reconstructions, and interpolations on CIFAR-10. Top three rows: samples, bottom three rows: reconstructions and interpolations. Our model achieves an Inception score of $6.88 (\pm 0.08)$, on par with the  $6.86 (\pm 0.06)$ achieved by Improved-GAN with historical averaging.}
\label{CIFAR_IAN}
\end{figure}

\begin{figure}[htbp]
\begin{center}
{\includegraphics[scale=0.85]{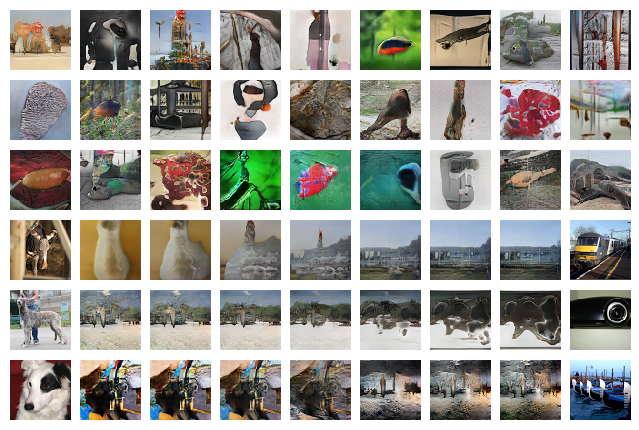}}
\end{center}
\caption{Samples, reconstructions, and interpolations on Imagenet. Top three rows: samples, bottom three rows: reconstructions and interpolations. Our model achieves an Inception score of $8.56 (\pm 0.09)$.}
\label{Imagenet}
\end{figure}

\end{document}